\documentclass[runningheads,a4paper]{llncs}

\usepackage{amssymb}
\usepackage{graphicx}

\usepackage{url}

\usepackage{todonotes}

\urldef{\mailsa}\path|{tsirizo.rabenoro, jerome.lacaille}@snecma.fr,|
\urldef{\mailsb}\path|{marie.cottrell, fabrice.rossi}@univ-paris1.fr|
\newcommand{\keywords}[1]{\par\addvspace\baselineskip
\noindent\keywordname\enspace\ignorespaces#1}
\usepackage[utf8]{inputenc}
\usepackage[english]{babel}
\usepackage{microtype}

\begin{document}

\mainmatter  

\title{A Methodology for the Diagnostic of Aircraft Engine Based on Indicators Aggregation}

\titlerunning{Diagnostic of Aircraft Engine Based on Indicators
Aggregation}

%
%
\author{Tsirizo Rabenoro\inst{1}%
\thanks{This study is supported by a grant from Snecma, Safran Group, one of the world’s leading manufacturers of aircraft and rocket engines, see
 \protect\url{http://www.snecma.com/} for details. }%
\and J\'{e}r\^{o}me Lacaille\inst{1} \and Marie Cottrell\inst{2}\and Fabrice Rossi\inst{2}}
\authorrunning{T. Rabenoro, J. Lacaille, M. Cottrell, F. Rossi}

\institute{Snecma, Groupe Safran,\\
77550 Moissy Cramayel, France\\
\and SAMM (EA 4543), Universit\'{e} Paris 1,\\
90, rue de Tolbiac, 75634 Paris Cedex 13, France\\
\email{\{tsirizo.rabenoro, jerome.lacaille\}@snecma.fr}\\
\email{\{marie.cottrell, fabrice.rossi\}@univ-paris1.fr}}

%
%

\toctitle{Diagnostic of Aircraft Engine Based on Indicators
Aggregation}
\tocauthor{T. Rabenoro, J. Lacaille, M. Cottrell, F. Rossi}
\maketitle

\begin{abstract}
  Aircraft engine manufacturers collect large amount of engine related data
  during flights. These data are used to detect anomalies in the engines in
  order to help companies optimize their maintenance costs. This article
  introduces and studies a generic methodology that allows one to build
  automatic early signs of anomaly detection in a way that is understandable by
  human operators who make the final maintenance decision. The main idea of
  the method is to generate a very large number of binary indicators based on
  parametric anomaly scores designed by experts, complemented by simple aggregations of
  those scores. The best indicators are selected via a classical forward
  scheme, leading to a much reduced number of indicators that are tuned to a
  data set. We illustrate the interest of the method on simulated data which
  contain realistic early signs of anomalies. 

\end{abstract}

\section{Introduction}
Aircraft engines are generally made extremely reliable by their conception
process and thus have low rate of operational events.  For example, in 2013,
the CFM56-7B engine, produced jointly by Snecma and GE aviation, has a
rate of in flight shut down (IFSD) is $0.02$ (per 1000 Engine Flight Hour) and
a rate of aborted take-off (ATO) is $ 0.005 $ (per 1000 departures).  This
dispatch availability of nearly 100 \% (99.962 \% in 2013) is obtained also
via regular maintenance operations but also via engine health monitoring (see
also e.g. \cite{vasov2007reliability} for an external evaluation). 

This monitoring is based, among other, on data transmitted by
satellites\footnote{using the commercial standard Aircraft Communications
  Addressing and Reporting System (ACARS, see
  \url{http://en.wikipedia.org/wiki/ACARS}), for instance.}  between aircraft
and ground stations.  Typical transmitted messages include engine status
overview as well as useful measurements collected as specific instants (e.g.,
during engine start). Flight after flight, measurements sent are analyzed in
order to detect anomalies that are early signs of degradation.  Potential
anomalies can be automatically detected by algorithms designed by experts.  If
an anomaly is confirmed by a human operator, a maintenance recommendation is
sent to the company operating the engine.

As a consequence, unscheduled inspections of the engine are sometimes
required. These inspections are due to the abnormal measurements. 
Missing a detection of early signs of degradation can result in an IFSD, an ATO or a delay and cancellation (D\&C). 
Despite the rarity of such events, companies need to avoid them to minimize
unexpected expenses and customers' disturbance. Even in cases where an
unscheduled inspection does not prevent the availability of the aircraft, it
has an attached cost: it is therefore important to avoid as much as possible
useless inspections.

We describe in this paper a general methodology to built complex automated
decision support algorithms in a way that is comprehensible by human
operators who take final decisions. The main idea of our approach is to leverage expert knowledge in
order to build hundreds of simple binary indicators that are all signs of the
possible existence of an early sign of anomaly in health monitoring data. The
most discriminative indicators are selected by a standard forward feature
selection algorithm. Then an automatic classifier is built on those
features. While the classifier decision is taken using a complex decision
rule, the interpretability of the features, their expert based nature and
their limited number allows the human operator to at least partially
understand how the decision is made. It is a requirement to have a trustworthy decision for the operator.

We will first describe the health monitoring context in Section
\ref{sec:context}. Then, we will introduce in more details the proposed
methodology in Section \ref{sec:arch-decis-proc}. Section
\ref{sec:simulation-study} will be dedicated to a simulation study that
validates our approach. 

\section{Context}\label{sec:context}
\subsection{Flight data}
\begin{figure}
  \begin{center}
    \includegraphics[width=0.6\linewidth]{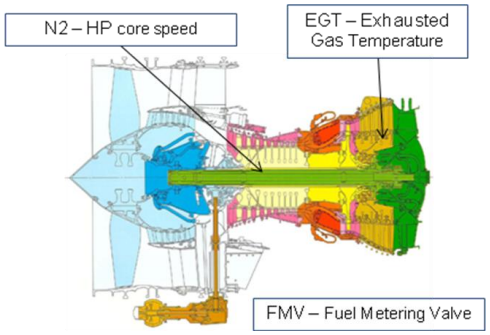}
  \end{center}
  \caption{Localization of some followed parameters on the Engine}
  \label{fig:startsequence}
\end{figure}

Engine health monitoring is based in part on flight data acquisition. Engines
are equipped with multiple sensors which measure different physical quantities
such as the high pressure core speed (N2), the Fuel Metering Valve (FMV),
the Exhausted Gas Temperature (EGT), etc. (See Figure \ref{fig:startsequence}.)
Those measures are monitored in real time during the flight. For instance the
quantities mentioned before (N2, FMV, etc.) are analyzed, among others,
during the engine starting sequence. This allows one to check the good health of the engine.
If potential anomaly is detected, a diagnostic is made. 
Based on the diagnostic sent to a company operator, the airline may have to
postpone the flight or cancel it, depending on the criticality of the fault
and the estimated repair time.

The monitoring can also be done flight after flight to detect any change
that can be flagged as early signs of degradations. Flight after
flight, measurements are compressed in order to obtain an overview of engines status 
that consists in useful measurements at specific recurrent moments. 
These useful measurements are then preprocessed to obtain measurements independent from external environment.
 These preprocessed data are analyzed by algorithms and human operators. The methodology
introduced in this article is mostly designed for this kind of monitoring.

\subsection{Detecting faults and abnormal behaviors}
Traditional engine health monitoring is strongly based on expert knowledge and field
experience (see e.g. \cite{tumer1999survey} for a survey and
\cite{flandrois2009expertise} for a concrete example). 
Faults and early signs of faults are
identified from suitable measurements associated to adapted computational
transformation of the data. For instance, the different measurements
(temperatures, vibration, etc.) are influenced by the flight parameters
(e.g. throttle position) and conditions (outside temperature,
etc.). Variations in the measured values can therefore result from variations
in the parameters and conditions rather than being due to abnormal behavior. Thus a
typical computational transformation consists in preprocessing the measurements
in order to remove dependency to the flight context
\cite{lacaille2009standardized}. 

In practice, the choice of measurements and computational transformations is
generally done based on expert knowledge.  For instance in
\cite{rabenoroinstants}, a software is designed to record expert decision
about a time interval on which to monitor the evolution of such a measurement
(or a time instant when such a measurement should be recorded). Based on the
recorded examples, the software calibrates a pattern recognition model that
can automatically reproduce the time segmentation done by the expert. Once the
indicators have been computed, the normal behavior of the indicators can be
learned. The residuals between predictions and actual indicators can be
statistically modeled, e.g. as a Gaussian vector. A score measurement is
obtained from the likelihood of this distribution. The normalized vector is a
failure score signature that may be described easily by experts to identify
the fault origin, in particular because the original indicators have some
meaning for them. See \cite{come2010aircraft}, \cite{flandrois2009expertise}
and \cite{lacaille2009maturation} for other examples.

However experts are generally specialized on a particular subsystem, thus each
algorithm focuses mainly on a specific subsystem despite the need of a
diagnostic of the whole system.

\subsection{Data and detection fusion}
The global diagnostic is currently done by the operator who collects all
available results of diagnostic applications. The task of taking a decision
based on all incoming information originating from different subsystems is
difficult. A first difficulty comes from dependencies between subsystems which
means that for instance in some situations, a global early sign of failure could be
detected by discovering discrepancies between seemingly perfectly normal
subsystems. In addition, subsystem algorithms can provide conflicting results
or a decision with a very low confidence level.  Furthermore, extreme
reliabilities of engines lead to an exacerbated trade off between false alarm
levels and detection levels, leading in general to a rather high level of
false alarms, at least at the operator level. Finally, the role of the
operator is not only to identify a possible early sign of failure, but also to
issue recommendations on the type of preventive maintenance needed. In other
words, the operator needs to identify the possible cause of the potential
failure. 

\subsection{Objectives}
The long term goal of engine health monitoring is to reach automated accurate,
trustworthy and precise maintenance decisions during optimally scheduled shop visit,
 but also to drastically reduce operational events such as IFSD and ATO. 
However, partly because of the current industrial standard,
 pure black box modeling is unacceptable. Indeed, operators are
currently trained to understand expertly designed indicators and to take
complex integrated decisions on their own. In order for a new methodology to
be accepted by operators, it has at least to be of a gray box nature, that is
to be (partially) explainable via logical and/or probabilistic
reasoning. Then, our objective is to design a monitoring methodology that
helps the human operator by proposing integrated decisions based on expertly
designed indicators with a ``proof of decision''.

\section{Architecture of the Decision Process}\label{sec:arch-decis-proc}
\subsection{Health monitoring data}
In order to present the proposed methodology, we first describe the data
obtained via health monitoring and the associated decision problem. 

We focus here on ground based long term engine health monitoring. Each flight
produces dozens of timestamped flight events and data. Concatenating those
data produces a multivariate temporal description of an engine whose
dimensions are heterogeneous. In addition, sampling rates of individual
dimensions might be different, depending on the sensors, the number of
critical time points recorded in a flight for said sensor, etc.

Based on expert knowledge, this complex set of time series is turned into a
very high dimensional indicator vector. The main idea, outlined in the
previous section, is that experts generally know what is the expected behavior
of a subsystem of the engine during each phase of the flight. Then the
dissimilarity between the expected behavior and the observed one can be
quantified leading to one (or several) anomaly scores. Such scores are in
turn transformed into binary indicators where $1$ means an anomaly is detected
and $0$ means no anomaly detected.

This transformation has two major advantages: it homogenizes the data and
it introduces simple but informative features (each indicator is associated to a
precise interpretation related to expert knowledge). It leads also to a loss
of information as the raw data are in general non recoverable from the
indicators. This is considered here a minor inconvenience as long as the
indicators capture all possible failure modes. This will be partially
guaranteed by including numerous variants of each indicator (as explained
below). On a longer term, our approach has to be coupled with field experience
feedback and expert validation of its coverage.

After the expert guided transformation, the monitoring problem becomes a rather
standard classification problem: based on the binary indicators, the decision
algorithm has to decide whether there is an anomaly in the engine and if, this
is the case, to identify the type of the anomaly (for instance by identifying
the subsystem responsible for the potential problem).

We describe now in more details the construction of the binary indicators. 

\subsection{Some types of anomalies}\label{sec:some-types-anomalies}
\begin{figure}
\begin{center}
\scriptsize
\begin{minipage}[c]{0.32\textwidth}
    \includegraphics[width=\linewidth]{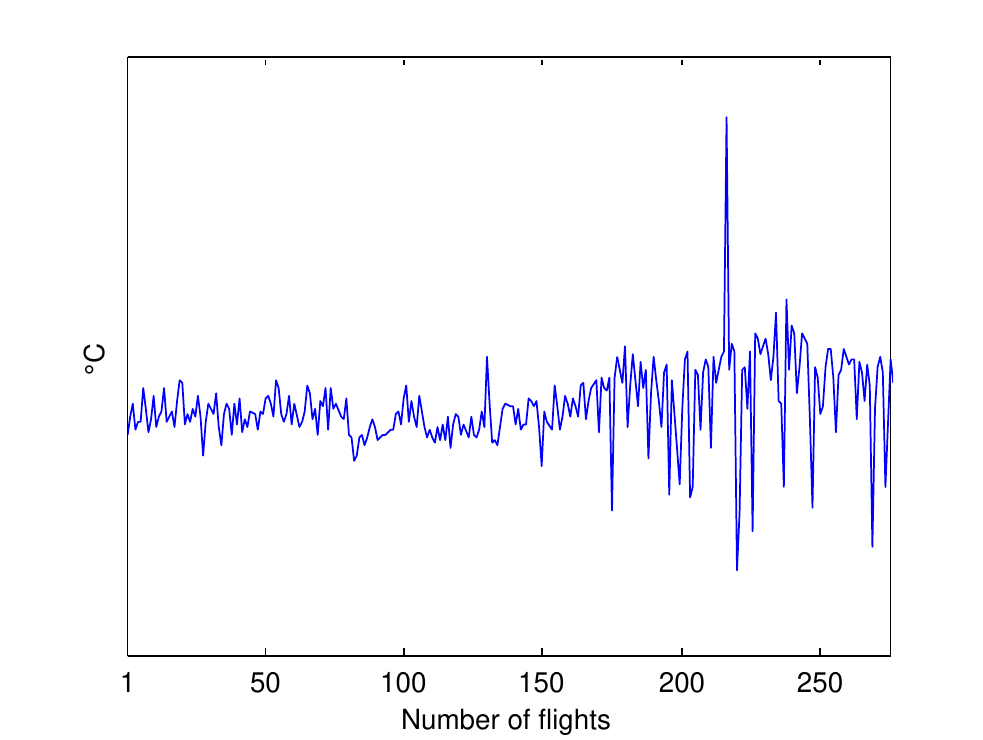}
    \caption{\scriptsize Variance shift}
   \label{fig:var1}
\end{minipage}
\begin{minipage}[c]{0.32\textwidth}
    \includegraphics[width=\linewidth]{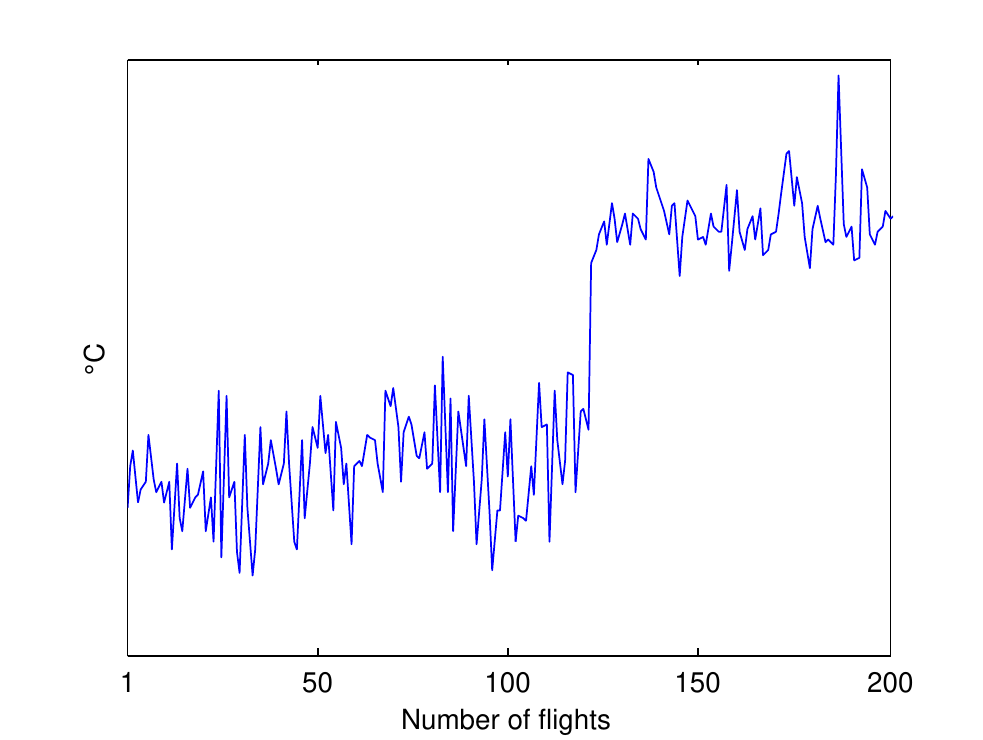}
    \caption{\scriptsize Mean shift}
   \label{jump1}
\end{minipage}
\begin{minipage}[c]{0.32\textwidth}
    \includegraphics[width=\linewidth]{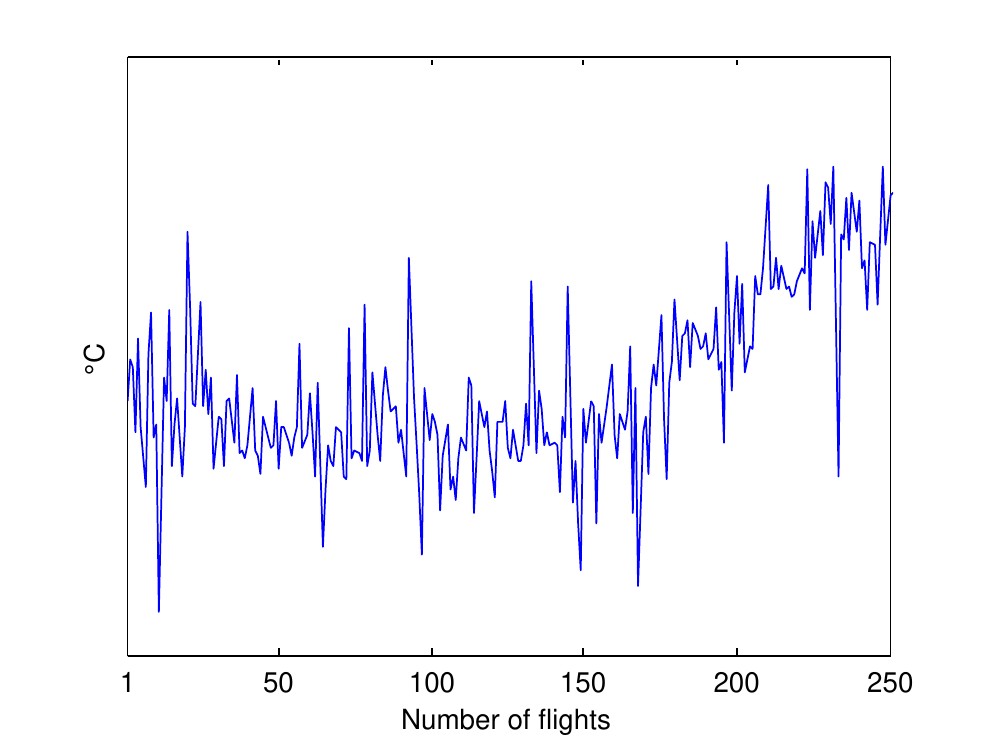}
    \caption{\scriptsize Trend modification}
   \label{trend1}
\end{minipage}
\normalsize
\end{center}
\end{figure}

Some typical univariate early signs of anomalies are shown on Figures \ref{fig:var1},
\ref{jump1} and \ref{trend1} which display the evolution through time of a
numerical value extracted from real world data. One can identify,
with some practice, a variance shift on Figure \ref{fig:var1}, a mean shift
on Figure \ref{jump1} and a trend modification (change of slope) on Figure
\ref{trend1}. In the three cases, the change instant is roughly at the center
of the time window. 

The main assumption used by experts in typical situations is that, when
external sources of change have been accounted for, the residual signal should
be stationary in a statistical sense. That is, observations $$\mathcal{Y}_n =
(Y_1(\theta_1), ..., Y_n(\theta_n))$$ are assumed to be generated identically
and independently from a fixed parametric law, with a constant set of
parameters (that is, all the $\theta_i$ are identical). Then, detecting an
anomaly amounts to detecting a change in the time series (as illustrated by
the three Figures above). This can be done via numerous well known statistical
tests \cite{basseville1995detection}. In the multivariate cases, similar
shifts in the signal can be associated to anomalies. More complex scenarios,
involving for instance time delays, can also be designed by experts. 

\subsection{From anomaly types to indicators}\label{sec:from-anomaly-types}
While experts can generally describe explicitly what type of change they are
expecting for some specific early signs of anomaly, they can seldom provide
detailed parameter settings for statistical tests (or even for the aggregation
technique that could lead to a statistical test after complex
calculations). To maximize coverage it seems natural to include numerous
indicators based on variations of the anomaly detectors compatible with expert
knowledge.

Let us consider for illustration purpose that the expert recommends to look
for shifts in mean of a certain quantity as early signs of a specific
anomaly. If the expert believes the quantity to be normally distributed with a
fixed variance, then a natural test would be Student's t-test. If the expert
has no strong priors on the distribution, a natural test would be the
Mann–Whitney U test. Both can be included to maximize coverage.

Then, in both cases, one has to assess the scale of the shift. Indeed those
tests work by comparing summary statistics of two populations, before and
after a possible change point. To define the populations, the expert has to
specify the length of the time windows to consider before and after the
possible change point: this is the expected scale at which the shift will
appear. In most cases, the experts can only give a rough idea of the
scale. Again, maximizing the coverage leads to the inclusion of several scales
compatible with the experts' recommendations. 

Given the choice of the test, of its scale and of a change point, one can
construct a statistic. A possible choice for the indicator could be this
value or the associated $p$-value. However, we choose to use simpler indicators
to ease their interpretation. Indeed, the raw value of a statistic is
generally difficult to interpret. A $p$-value is easier to understand because
of the uniform scale, but can still lead to misinterpretation by operators
with insufficient statistical training. We therefore choose to use binary
indicators for which the value 1 corresponds to a rejection of the null
hypothesis of the underlying test to a given level (the null hypothesis is
here the case with no mean shift). 

Finally, as pointed out before, aircraft engines are extremely reliable, a fact
that increases the difficulty in balancing sensibility and specificity of anomaly
detectors. In order to alleviate this difficulty, we build high level
indicators from low level tests. For instance, if we monitor the evolution of
a quantity on a long period compared to the expected time scale of anomalies,
we can compare the number of times the null hypothesis of a test has been
rejected on the long period with the number of times it was not rejected, and
turn this into a binary indicator with a majority rule. 

To summarize, we construct parametric anomaly scores from expert knowledge,
together with acceptable parameter ranges. By exploring those ranges, we
generate numerous (possible hundreds of) binary indicators. Each indicator can be
linked to an expertly designed score with a specific set of parameters and
thus is supposedly easy to interpret by operators. Notice that while we as
focused in this presentation on temporal data, this framework can be applied
to any data source. 

\subsection{Decision}
The final decision step consists in classifying these high dimensional binary
vectors into at least two classes, i.e., the presence or absence of an
anomaly. A classification into more classes is highly desirable if possible,
for instance to further discriminate between seriousness of anomalies and/or
sources (in terms of subsystems of the engine). In this paper however, we will
restrict ourselves to a binary classification case (with or without anomaly).

As explained before, we aim in the long term at gray box modeling, so while
numerous classification algorithms are available see
e.g. \cite{kotsiantis2007supervised}, we shall focus on interpretable ones. In
this paper, we choose to use Random Forests \cite{breiman2001random} as they
are very adapted to binary indicators and to high dimensional data. They are
also known to be robust and to provide state-of-the-art classification
performances at a very small computational cost. While they are not as
interpretable as their ancestors CART \cite{breiman1984classification}, they
provide at least variable importance measures that can be used to identify the
most important indicators. 

Finally, while including hundreds of indicators is important to give a broad
coverage of the parameter spaces of the expert scores and thus to maximize the
probability of detecting anomalies, it seems obvious that some redundancy will
appear. Therefore, we have chosen to apply a feature selection technique
\cite{guyon2003introduction} to this problem. The reduction of number of
features will ease the interpretation by limiting the quantity of information
transmitted to the operators in case of a detection by the classifier. Among
the possible solutions, we choose to use the Mutual information based
technique Minimum Redundancy Maximum Relevance (mRMR, \cite{peng2005feature})
which was reported to give excellent results on high dimensional data.

\section{A simulation study}\label{sec:simulation-study}

\subsection{Introduction}
It is difficult to find real data with early signs of degradations, 
because their are scarce and moreover the scheduled maintenance operations tend to remove
these early signs.  Experts could study in detail recorded data to find early
signs of anomalies whose origins were fixed during maintenance but it is close
to looking for a needle in a haystack, especially considering the huge amount
of data to analyze. We will therefore rely in this paper on
simulated data. Our goal is to validate the interest of the proposed
methodology in order to justify investing in the production of carefully
labelled real world data. 

In this section we begin by the description of the simulated data used for the
evaluation of the methodology, and then we will present the performance
obtained on this data.

\subsection{Simulated data}
The simulated data are generated according to the univariate shift models
described in Section \ref{sec:some-types-anomalies}. We generate two data sets
a simple one $A$ and a more complex one $B$. 

In the first case $A$, it is assumed that expert based normalisation has been
performed. Therefore when no shift in the data distribution occurs, we observe
a stationary random noise modeled by the standard Gaussian distribution. Using
notations of Section \ref{sec:some-types-anomalies} the $Y_i$ are independent
and identically distributed according to
$\mathcal{N}(\mu=0,\sigma^2=1)$. Signals in set $A$ have a length chosen
uniformly at random between 100 and 200 observations (each signal has a
specific length).

\begin{figure}[htbp]
\begin{center}
\scriptsize
\begin{minipage}[c]{0.45\textwidth}
    \includegraphics[width=\linewidth,height=0.575\linewidth]{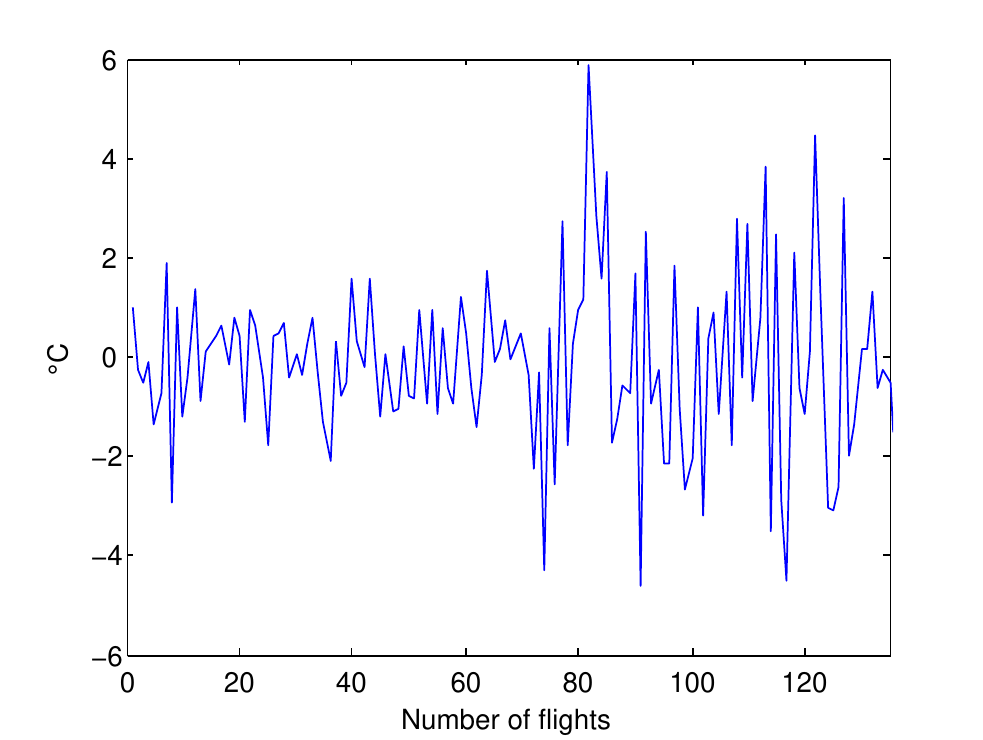}
    \caption{\scriptsize Variance shift}
   \label{Fig:var2}
\end{minipage}
\begin{minipage}[c]{0.45\textwidth}
    \includegraphics[width=\linewidth,height=0.575\linewidth]{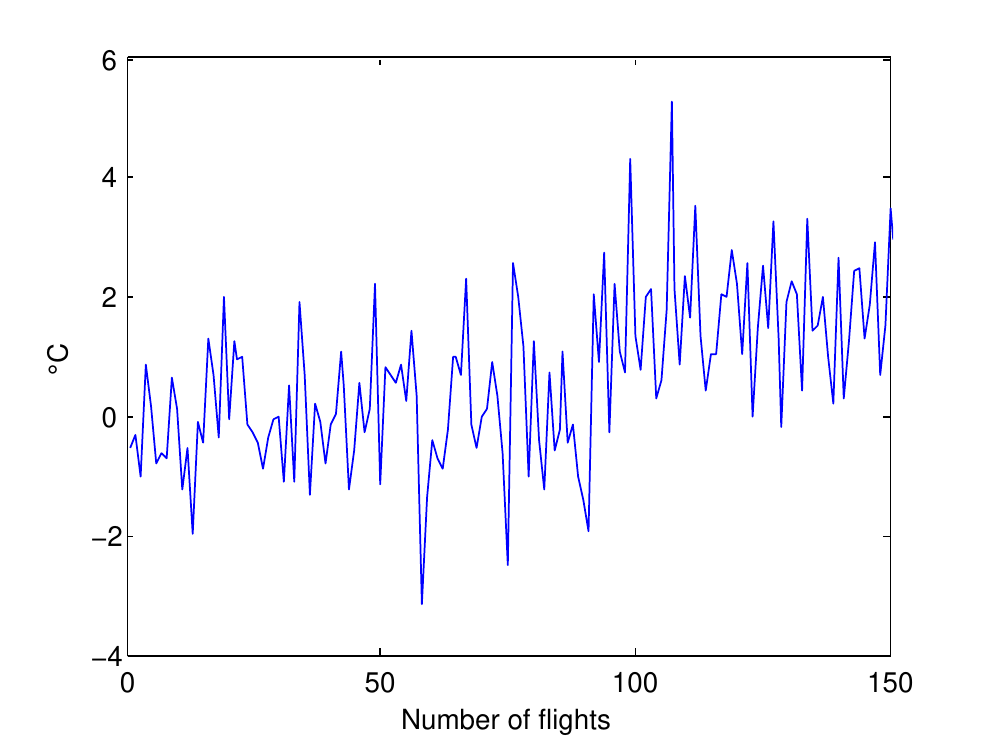}
    \caption{\scriptsize Mean shift}
   \label{Fig:jump2}
\end{minipage}

\begin{minipage}[t]{0.45\textwidth}
    \includegraphics[width=\linewidth,height=0.575\linewidth]{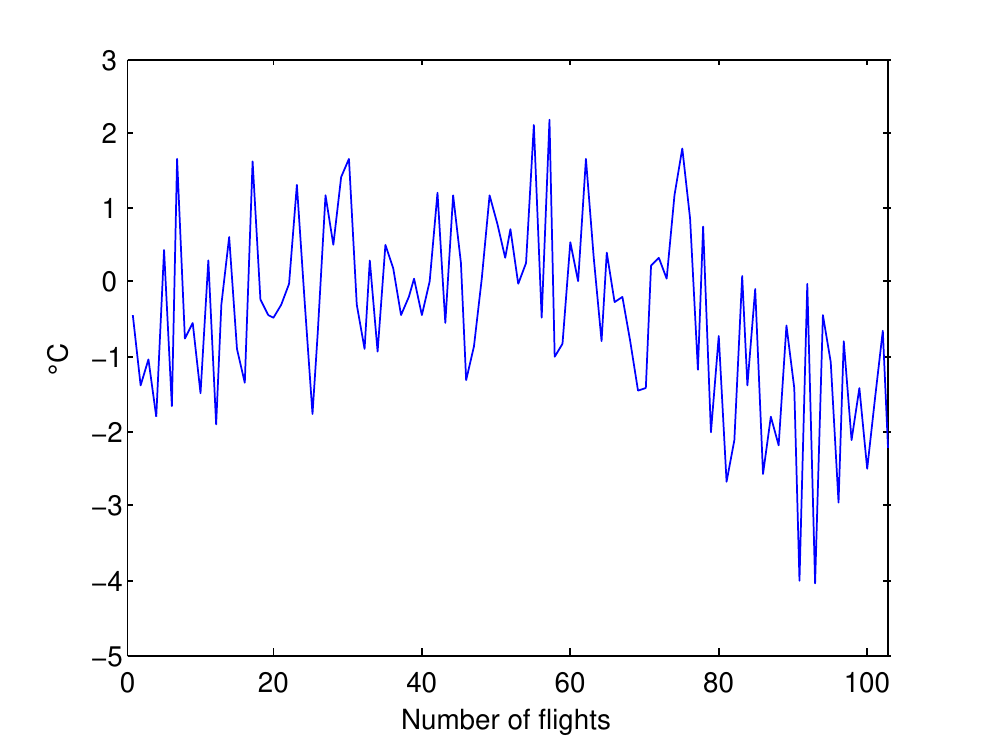}
    \caption{\scriptsize Trend modification}
   \label{Fig:trend2}
\end{minipage}
\begin{minipage}[t]{0.45\textwidth}
    \includegraphics[width=\linewidth,height=0.575\linewidth]{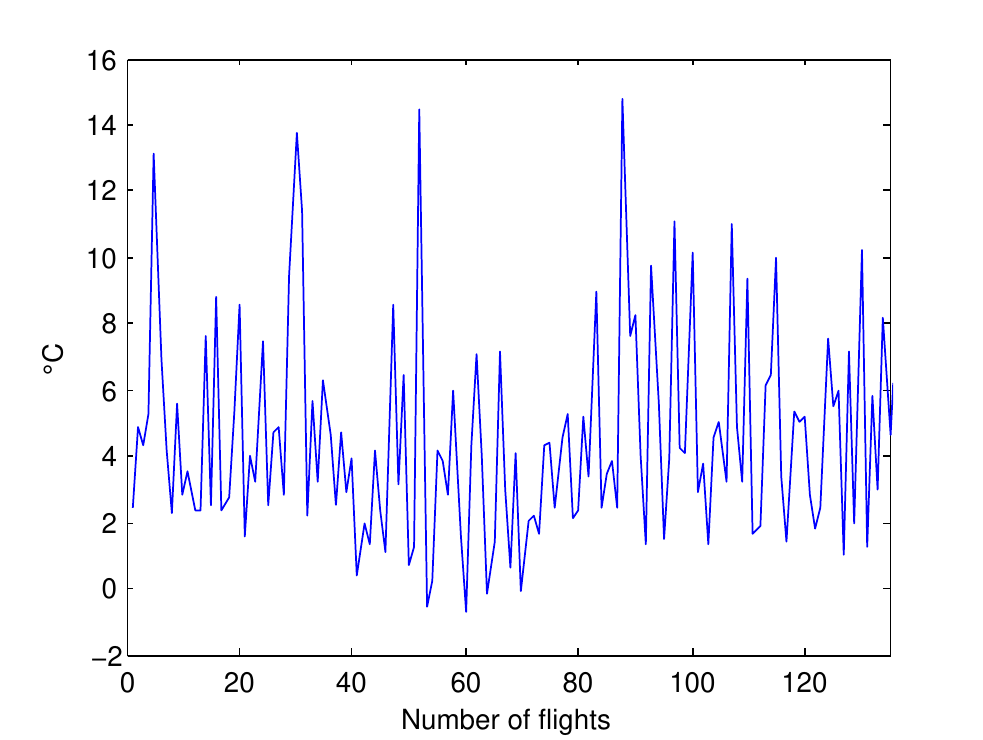}
    \caption{Data without anomaly but with suboptimal normalisation
      represented by a slow varying deterministic component}
   \label{sin1}
 \end{minipage}
 \normalsize
\end{center}
\end{figure}

Anomalies are modelled after the three examples given in Figures
\ref{fig:var1}, \ref{jump1} and \ref{trend1}. We implement therefore three
types of shift: 
\begin{enumerate}
\item a variance shift: in this case,
  observations are distributed according to
  $\mathcal{N}(\mu=0,\sigma^2)$ with $\sigma^2=1$ before the change point and $\sigma$ chosen uniformly
  at random in $[1.01, 5]$ after the change point (see Figure \ref{Fig:var2});
\item a mean shift: in this case,
  observations are distributed according to
  $\mathcal{N}(\mu,\sigma^2=1)$ with $\mu=0$ before the change point and $\mu$ chosen uniformly
  at random in $[1.01, 5]$ after the change point (see Figure
  \ref{Fig:jump2});
\item a slope shift: in this case , observations are distributed according to
  $\mathcal{N}(\mu,\sigma^2=1)$ with $\mu=0$ before the change point and $\mu$
  increasing linearly from $0$ from the change point with a slope chosen uniformly
  at random in $[0.02,3]$ (see Figure
  \ref{Fig:trend2}).
\end{enumerate}
Assume that the signal contains $n$ observations, then the change point is
chosen uniformly at random between the $\frac{2n}{10}$-th observation and the
$\frac{8n}{10}$-th observation. 

According to this procedure, we generate a balanced data set with 6000
observations corresponding to 3000 observations with no anomaly, and 1000
observations for each of the three types of anomalies.

In the second data set, $B$, a slow deterministic variation is added to
randomly chosen signals with no anomaly: this is a way to simulate a
suboptimal normalisation (see Figure \ref{sin1} for an example). The slow
variation is implemented by adding to the base noise a sinus with a period
of $\frac{2}{3}$ of the signal length and amplitude 1.

Signals in set $B$ are shorter, to make the detection more difficult: they are chosen
uniformly at random between 100 and 150 observations. In addition, the noise 
is modeled by a $\chi^2$ distribution with 4 degrees of freedom. Signals with
an anomaly are generated using the same rationale as for set $A$. In this case
however, the mean shift is simply implemented by adding a constant to the
signal after the change point. The ``variance'' shift is in fact a
change in the number of degrees of freedom of the $\chi^2$ distribution: after
the change point, the number of degrees is chosen randomly (uniformly) between
8 and 16. The change point is chosen as in set $A$.

According to this procedure, we generate a balanced data set with 6000
observations corresponding to 3000 observations with no anomaly, and 1000
observations for each of the three types of anomalies. Among the 3000 anomaly
free signals, 1200 are corrupted by a slow variation. 

\subsection{Indicators}
As explained in Section \ref{sec:from-anomaly-types}, binary indicators are
constructed from expert knowledge by varying parameters, including scale and
position parameters. In the present context, we use sliding windows: for each
position of the window, a classical statistical test is conducted to decide
whether a shift in the signal occurs at the center of the window. 

The ``expert'' designed tests are here:
\begin{enumerate}
\item the Mann–Whitney–Wilcoxon U test (non parametric test for shift in
  mean);
\item the two sample Kolmogorov-Smirnov test (non parametric test for
  differences in distributions);
\item the F-test for equality of variance (parametric test based on a Gaussian
  hypothesis). 
\end{enumerate}
The direct parameters of those tests are the size of the window which defines
the two samples (30, 50, and $\min(n-2,100)$ where $n$ is the signal length) and
the level of significance of the test (0.005, 0.1 and 0.5). Notice that those
tests do not include a slope shift detection. 

Then, more complex binary indicators are generated, as explained in Section
\ref{sec:from-anomaly-types}. In a way, this corresponds to build very simple
binary classifiers. We use the following ones:
\begin{enumerate}
\item for each underlying test, the derived binary indicator takes the value
  one if on a fraction $\beta$ of $m$ windows, the test detects a change. 
  Parameters are the test itself with its parameters, the value of $\beta$
  (we considered 0.1, 0.3 and 0.5) and the number of observations in common
  between two consecutive windows (the length of the window minus 1, 5 or
  10);
\item for each underlying test, the derived binary indicator takes the value
  one if on a fraction $\beta$ of $m$ consecutive windows, the test detects a
  change (same parameters);
\item for each underlying test, the derived binary indicator takes the value
  one if there are 5 consecutive windows such that the test detects a change
  on at least $k$ of these 5 consecutive windows (similar parameters where $\beta$ is replaced by $k$).
\end{enumerate}
In addition, based on expert recommendation, we apply all those indicators
both to the original signal and to a smoothed signal (using a simple moving
average over 5 measurements). 

We use more than 50 different configurations for each indicator, leading to a
total number of 810 binary indicators (it should be noted that only a subset
of all possible configurations is included into this indicator vector). 

\subsection{Reference performances}
In this paper, we focus on the simple case of learning to discriminate between
a stationary signal and a signal with a shift. We report therefore the
classification rate (classification accuracy). 

For both sets $A$ et $B$, the learning sample is composed of 1000 signals
keeping the balance between the three classes of shifts. The evaluation is
done on the remaining 5000 signals that have been divided in 10 groups of 500
time series each. The rationale of this data splitting in the evaluation phase
is to estimate both the prediction quality of the model but also the
variability in this rate as an indication of the trust we can put on the
results. We also use and report the out-of-bag (OOB) estimate of the
performances provided by the Random Forest (this is a byproduct of the
bootstrap procedure used to construct the forest, see
\cite{breiman2001random}). 

When all the 810 indicators are used, the classification performances are very
high on set $A$ and acceptable on set $B$ (see Table \ref{tab:full}). The
similarity between the OOB estimate of the performances and the actual
performances confirms that the OOB performances can be trusted as a reliable
estimator of the actual performances. Data set $B$ shows a strong over fitting
of the Random Forest, whereas data set $A$ exhibits a mild one.

\begin{table}[htbp]
  \centering
  \begin{tabular}{lccc}
Data set & \ Training set accuracy \ & OOB accuracy & \ Test set average accuracy \\\hline
$A$ & 1 & 0.953 &0.957 (0.0089) \\
$B$ & 1 & 0.828 &0.801 (0.032)\\\hline   
  \end{tabular}
\medskip
  \caption{Classification accuracy of the Random Forest using the 810 binary
    indicators. For the test set, we report the average classification
    accuracy and its standard deviation between parenthesis.}
  \label{tab:full}
\end{table}

\subsection{Feature selection}
On the data set $A$, the performances are very satisfactory, but the model is
close to a black box in the sense that it uses all the 810 indicators. Random
Forests are generally difficult to interpret, but a reduction in the number of
indicators would allow an operator to study the individual decisions performed
by those indicators in order to have a rough idea on how the global decision could
have been made. On the data $B$ set, the strong over fitting is another
argument for reducing the number of features.

Using the mRMR we ranked the 810 indicators according to a mutual information
based estimation of their predictive performances. We use then a forward
approach to evaluate how many indicators are needed to achieve acceptable
predictive performances. Notice that in the forward approach, indicators are
added in the order given by mRMR and then never removed. As mRMR takes into
account redundancy between the indicators, this should not be a major
issue. Then for each number of indicators, we learn a Random Forest on the
learning set and evaluate it. 

\begin{figure}
\centering
\includegraphics[width=0.8\linewidth]{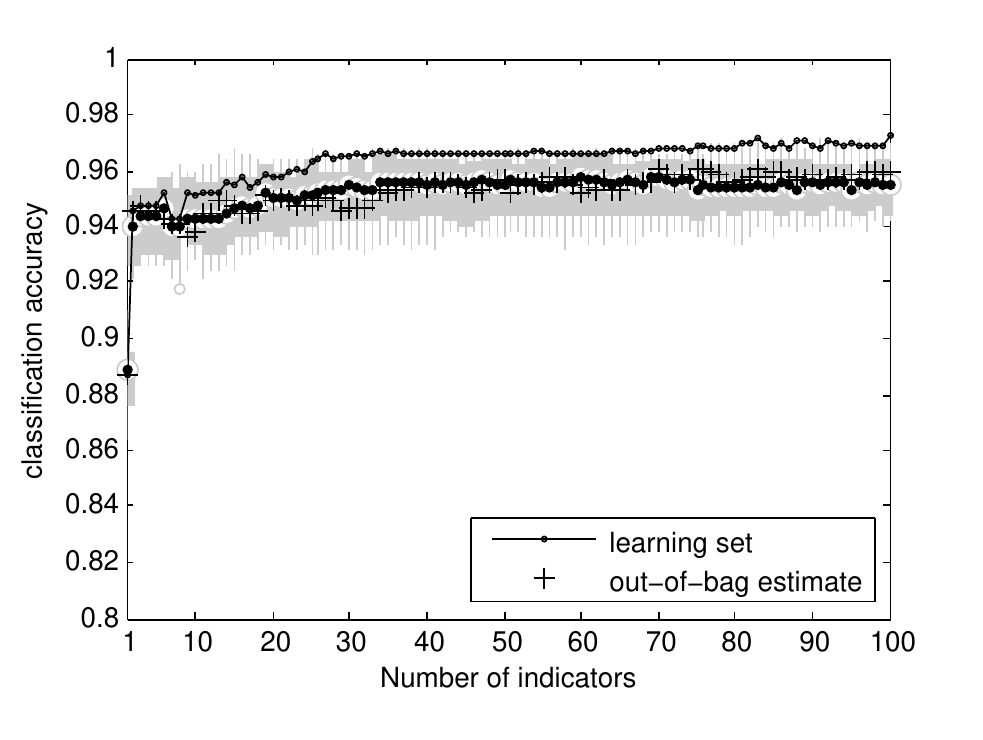}
\caption{\textbf{Data set $A$} : classification accuracy on learning set (circle) as a function of the
  number of indicators. A boxplot gives the classification accuracies on the
  test subsets, summarized by its median (black dot inside a white
  circle). The estimation of those accuracies by the out-of-bag (OOB) bootstrap
  estimate is shown by the crosses.}
\label{fig:box}
\end{figure}

Figure \ref{fig:box} shows the results for data set $A$. Accuracies are
quite high with a rather low number of indicators, with a constant increase in
performances on the learning set (as expected) and a stabilisation of the real
performances (as evaluated by the test set and the OOB estimation) around
roughly 40 indicators. 

\begin{figure}
\centering
\includegraphics[width=0.8\linewidth]{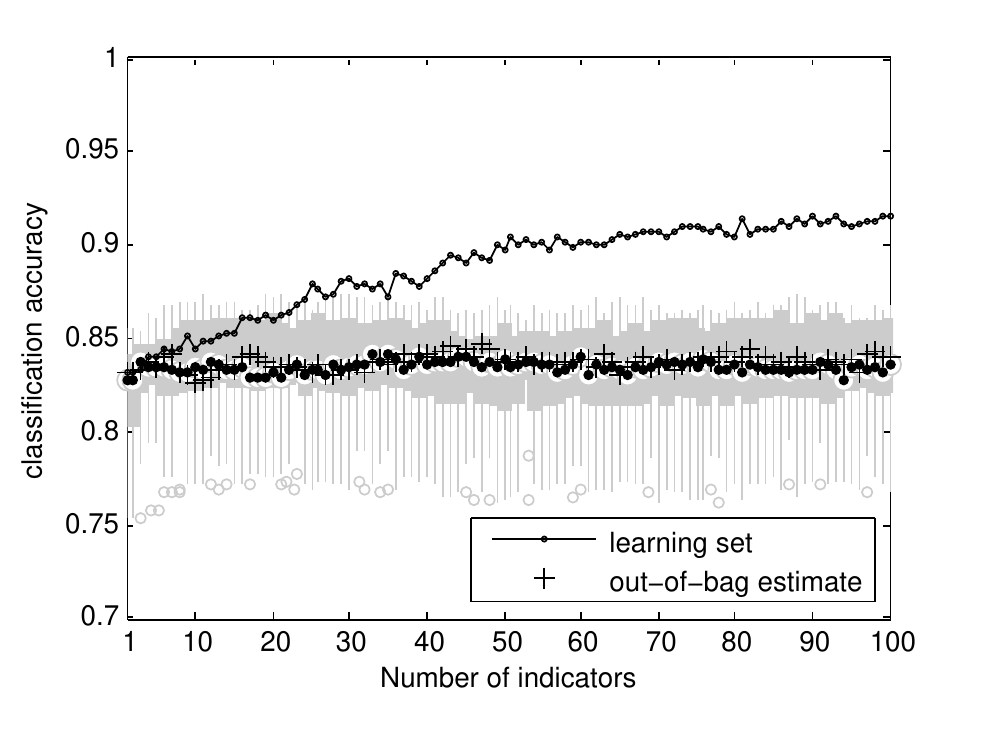}
\caption{\textbf{Data set $B$} : classification accuracy on learning set (circle) as a function of the
  number of indicators. A boxplot gives the classification accuracies on the
  test subsets, summarized by its median (black dot inside a white
  circle). The estimation of those accuracies by the out-of-bag (OOB) bootstrap
  estimate is shown by the crosses.}
\label{fig:ctype2}
\end{figure}

Figure \ref{fig:ctype2} shows the results for data set $B$. Excepted the lower
performances and the stronger over fitting, the general behavior is similar to
the case of data set $A$. Those lower performances were expected because
data set $B$ has been designed to be more difficult to analyze, in part
because of the inadequacy between the actual data model ($\chi^2$
distribution) and the ``expert'' based low level tests (in particular the F
test which assumes Gaussian distributions).

In both cases, the feature selection procedure shows that only a small subset
of the original 810 indicators is needed to achieve the best performances
reachable on the data sets. This is very satisfactory as this allows to
present to operators a manageable number of binary decisions together with the
aggregated one provided by the random forest.  

\subsection{Selected indicators}
In order to illustrate further the interest of the proposed methodology, we
show in Table \ref{tab:tenbest:A} the best ten indicators for data set
$A$. Those indicators lead to quite good performances with an average test set
classification accuracy of 0.944 (OOB estimation is 0.938). 

\begin{table} 
  \begin{center}
    \begin{tabular}{ccccc}
      type of indicator & level & window length& window step \\\hline
      F test&0.005&100&5\\
      confu(2,3) &0.005&50&5\\
      ratef(0.1)&0.005&50&5\\
      KS test&0.005&100&1\\
      conff(3,5)&0.005&100&5\\
      KS test&0.1&100&5\\
      F test&0.005&100&1\\
      KS test&0.005&100&10\\
      lseqf(0.1)&0.1&50&1\\
      F test&0.005&50&10\\\hline
    \end{tabular}
  \end{center}
\caption{The best ten indicators according to mRMR for data set
  $A$. Confu(k,n) corresponds to a positive Mann–Whitney–Wilcoxon U test on
  k windows out of n consecutive ones. Conff(k,n) is the same thing for the
  F-test. Ratef($\alpha$) corresponds to a positive F-test on $\alpha\times m$
windows out of $m$. Lseqf($\alpha$) corresponds to a positive F-test on $\alpha\times m$
consecutive windows out of $m$. Lsequ($\alpha$) is the same for a U test. Here, none of the indicators are based on a
smoothed version of the signal.}
\label{tab:tenbest:A}
\end{table}

Table \ref{tab:tenbest:B} shows the best ten indicators for data set
$B$. Again, this corresponds to quite good performances with an average test set
classification accuracy of 0.831 (OOB estimation is 0.826). As
expected, the F test and indicators based on it are less interesting for this
data set as the noise is no more Gaussian. 
\begin{table} 
  \begin{center}
    \begin{tabular}{ccccc}
      type of indicator & level & window length& smoothed & window step \\\hline
KS test&0.005&100&no&5\\
lseqf(0.1)&0.1&30&yes&1\\
confu(4,5)&0.005&30&no&1\\
U test&0.1&100&no&5\\
confu(4,5)&0.005&100&no&5\\
confu(2,3)&0.005&100&no&1\\
lsequ(0.3)&0.1&50&no&1\\
F test &0.005&100&yes&10\\
confu(2,3)&0.005&30&no&5\\
KS Test&0.005&100&no&10\\\hline
\end{tabular}
\end{center}
\caption{The best ten indicators according to mRMR for data set
  $B$. Please refer Table \ref{tab:tenbest:A} for notations. }
\label{tab:tenbest:B}
\end{table}

In both cases, we see that the feature selection method is able to make a
complex selection in a very large set of binary indicators. This induces
indirectly an automatic tuning of the parameters of the low level tests and of
simple aggregation classifiers. Because of their simplicity and their binary
outputs, indicators are easy to understand by an operator.

\section{Conclusion and perspectives}
In this paper, we have introduced a diagnostic methodology for engine health
monitoring that leverage expert knowledge and automatic classification. The
main idea is to build from expert knowledge parametric anomaly scores
associated to range of plausible parameters. From those scores, hundreds of
binary indicators are generated in a way that covers the parameter space as
well as introduces simple aggregation based classifiers. This turns the
diagnostic problem into a classification problem with a very high number of
binary features. Using a feature selection technique, one can reduce the
number of useful indicators to a humanly manageable number. This allows a
human operator to understand at least partially how a decision is reached by
an automatic classifier. This is favored by the choice of the indicators which
are based on expert knowledge and on very simple decision rules. A very
interesting byproduct of the methodology is that is can work on very
different original data as long as expert decision can be modelled by a set of
parametric anomaly scores. This was illustrated by working on signals of
different lengths.  

Using simulated data, we have shown that the methodology is sound: it reaches
good predictive performances even with a limited number of indicators (e.g.,
10). In addition, the selection process behaves as expected, for instance by
discarding statistical tests that are based on hypothesis not fulfilled by the
data. However, we limited ourselves to univariate data and to a binary
classification setting (i.e., abnormal versus normal signal). We need to show
that the obtained results can be extended to multivariate data and to complex
classification settings (as identifying the cause of a possible anomaly is
extremely important in practice). 

It should also be noted that we relied on Random Forests which are not as easy
to interpret as other classifiers (such as CART). In our future work, we will
compare Random Forest to simpler classifiers. As we are using binary
indicators, some form of majority voting is probably the simplest possible
rule but using such as rule implies to choose very carefully the indicators
\cite{ruta2005classifier}. 

Finally, it is important to notice that the classification accuracy is
not the best way of evaluating the performances of a classifier in the health
monitoring context. Firstly, health monitoring intrinsically involves a strong
class imbalance \cite{japkowicz2002class}. Secondly, health monitoring is a
cost sensitive area because of the strong impact on airline profit of an
unscheduled maintenance. It is therefore important to take into account
specific asymmetric misclassification cost to get a proper performance
evaluation.



\bibliography{Biblio}
\bibliographystyle{splncs03}


\end{document}